\documentclass[sigconf]{acmart}

\AtBeginDocument{%
  \providecommand\BibTeX{{%
    \normalfont B\kern-0.5em{\scshape i\kern-0.25em b}\kern-0.8em\TeX}}}

\usepackage{graphicx}
\usepackage[spanish,ngerman,main=english]{babel}
\usepackage[utf8]{inputenc}
\usepackage{amsfonts}
\usepackage{amsmath}
\newcommand\norm[1]{\left\lVert#1\right\rVert}
\usepackage{algorithmic}
\usepackage[linesnumbered,ruled,vlined]{algorithm2e}

\SetCommentSty{mycommfont}
\SetKwInput{KwInput}{Input}                
\SetKwInput{KwOutput}{Output}              
\SetKwInput{KwRequire}{Require}              
\usepackage{array}
\usepackage{soul}
\usepackage{multirow}
\usepackage{xtab,booktabs,rotating}
\usepackage[caption=true,font=footnotesize]{subfig}
\usepackage{xhfill}
\usepackage{caption}
\usepackage[export]{adjustbox}
\usepackage{makecell}
\usepackage{url}
\usepackage{hyperref}
\usepackage[colorinlistoftodos]{todonotes}
\usepackage{microtype}        

\newcommand{\yd}[1]{{\color{black} #1}}

\renewcommand\footnotetextcopyrightpermission[1]{}

\acmConference[IPSN '22]{The International Conference on Information Processing in Sensor Networks}{May 4--6, 2022}{Milan, Italy (Virtual)}

\begin{document}

\title{YONO: Modeling Multiple Heterogeneous Neural Networks on Microcontrollers}

\author{Young D. Kwon}
\affiliation{%
  \institution{University of Cambridge}
  \country{United Kingdom}
}
\email{ydk21@cam.ac.uk}

\author{Jagmohan Chauhan}
\affiliation{%
  \institution{University of Southampton}
  \country{United Kingdom}
}
\email{J.Chauhan@soton.ac.uk}

\author{Cecilia Mascolo}
\affiliation{%
  \institution{University of Cambridge}
  \country{United Kingdom}
}
\email{cm542@cam.ac.uk}

\renewcommand{\shortauthors}{Kwon et al.}

\begin{abstract}
    Internet of Things (IoT) systems provide large amounts of data on all aspects of human behavior. Machine learning techniques, especially deep neural networks (DNN), have shown promise in making sense of this data at a large scale. Also, the research community has worked to reduce the computational and resource demands of DNN to compute on low-resourced microcontrollers (MCUs). However, most of the current work in embedded deep learning focuses on solving a single task efficiently, while the multi-tasking nature and applications of IoT devices demand systems that can handle a diverse range of tasks (such as activity, gesture, voice, and context recognition) with input from a variety of sensors, simultaneously.

    
    In this paper, we propose YONO, a product quantization (PQ) based approach that compresses multiple heterogeneous models and enables in-memory model execution and model switching for dissimilar multi-task learning on MCUs. We first adopt PQ to learn codebooks that store weights of different models. Also, we propose a novel network optimization and heuristics to maximize the compression rate and minimize the accuracy loss. Then, we develop an online component of YONO for efficient model execution and switching between multiple tasks on an MCU at run time without relying on an external storage device.

    YONO shows remarkable performance as it can compress multiple heterogeneous models with negligible or no loss of accuracy up to 12.37$\times$. Furthermore, YONO's online component enables an efficient execution (latency of 16-159 ms and energy consumption of 3.8-37.9 mJ per operation) and reduces model loading/switching latency and energy consumption by 93.3-94.5\% and 93.9-95.0\%, respectively, compared to external storage access. Interestingly, YONO can compress various architectures trained with datasets that were not shown during YONO's offline codebook learning phase showing the generalizability of our method. To summarize, YONO shows great potential and opens further doors to enable multi-task learning systems on extremely resource-constrained devices.
    


\end{abstract}




\maketitle

\section{Introduction}\label{sec:introduction}

With the rise of mobile, wearable devices, and the Internet of Things (IoT), the proliferation of sensory type data has fostered the adoption of deep neural networks (DNN) in the modeling of a variety of mobile sensing applications~\cite{lane_survey_2010}; researchers use DNN trained on sensory data in mobile sensing tasks such as human activity recognition~\cite{guan_ensembles_2017,wang_deep_2019}, gesture recognition~\cite{fan_what_2018}, tracking and localization~\cite{jiang_ariel:_2012}, mental health and wellbeing~\cite{lu_stresssense:_2012}, and audio sensing applications~\cite{purwins_deep_2019}.
While machine learning (ML) models are becoming more efficient on resource-constrained IoT devices~\cite{robert_tflm_mlsys2021}, most existing on-device systems, designed for microcontroller units (MCUs), are targeted at one specific application~\cite{banbury_micronets_2021,fedorov_sparse_2019,zhang_hello_2017}. Conversely, multi-application systems capable of directly supporting a wide range of applications on-device could be more versatile and useful in practice. Specifically, we envisage a system powered by MCUs that can recognize users' voice commands, activities and gestures, identify everyday objects and people, and understand the surrounding environments: this has the potential to boost the utilization of IoT devices in practice (e.g., help visually impaired individuals understand their environments~\cite{noauthor_wearable_2016}).


However, realizing such multi-tasking system faces three major challenges.
\textbf{First}, multiple dissimilar tasks based on different modalities of incoming data (e.g., voice recognition (audio), activity recognition (accelerometer signals), object classification (image)) need to co-exist in the same framework. As discussed in~\cite{lee_fast_2020}, conventional multi-task learning (MTL) approaches cannot address \textit{multiple heterogeneous networks} effectively. \textbf{Second}, IoT devices based on MCUs are extremely resource-constrained~\cite{fang_nestdnn:_2018,kusupati_fastgrnn:_2018}.
For example, ``high-end" MCUs (e.g., STMF767ZI) have only 512 KB Static Random-Access Memory (SRAM) for intermediate data and 2 MB on-chip embedded flash (eFlash) memory for program storage. 
\textbf{Finally}, in real-world deployment scenarios, context switching of different ML tasks at run-time could incur overheads on memory-constrained MTL systems as demonstrated in~\cite{lee_fast_2020}, where some models must reside in external storage devices due to the limited on-chip memory space. 
As on-chip memory operations are faster than external disk accesses, frequent model loading/swap between different tasks based on external storage increase the overall latency, exacerbating the usability and responsiveness of the system.

To solve these challenges, one of the common techniques employed is to compress individual models separately using pruning~\cite{han_deep_2016,zhang_systematic_2018} and quantization~\cite{jacob_quantization_2018}. However, model compression techniques are limited since extensive and iterative finetuning is required to ensure high performance after compressing a model. Also, since models are trained independently, they cannot benefit from potential knowledge transfer between different tasks. In the literature, researchers proposed MTL-based approaches to achieve robustness and generalization of multiple tasks, while increasing the compression rate of the model by sharing network structures. However, \textit{sharing/compressing multiple heterogeneous networks} has not been fully examined. 
Furthermore, prior work~\cite{lee_fast_2020} attempts to solve the MTL of multiple heterogeneous networks by sharing weights of multiple models via virtualization. However, this method is complex, and the compression ratio is constrained to 8.08$\times$ (see \S\ref{subsec:two modality} for detail), thereby limiting the type of IoT devices on which it can operate. Further, since only a simplified LeNet architecture is evaluated on an MCU, the system could not achieve high accuracy to be useful in practice (e.g., 59.26\% on the CIFAR-10 dataset~\cite{cifar10}).


\textbf{This Work.} To address the challenges and limitations of previous approaches, we propose \textbf{YONO} (\textbf{Y}ou \textbf{O}nly \textbf{N}eed \textbf{O}ne pair of codebooks), that adopts Product Quantization (PQ)~\cite{jegou_product_2011} to maximize compression rate and on-chip memory operations to minimize external disk accesses for heterogeneous multi-task learning. PQ, originally proposed in the database community, aims to decompose the original high-dimensional space into the Cartesian product of a finite number of low-dimensional subspaces that are independently quantized. A model's weight matrix of any layer can be converted to codeword indexes corresponding to the subvectors of the weight matrix via a codebook. 

Inspired by successful applications of PQ on approximate nearest neighbor search out of billions of vectors in the database community~\cite{jegou_product_2011,kalantidis_locally_2014,ge_optimized_2014} and single layer compression in an individual model~\cite{gong_compressing_2014,wu_quantized_2016,stock_and_2019,stock_training_2020,martinez_permute_2021}, we jointly apply PQ on multiple models instead of on a layer of a model. \yd{We find just one pair of codebooks that are generalizable and thus can be shared across many dissimilar tasks.} We then propose a novel optimization process based on alternating PQ and finetuning steps to mirror the performance of the original models. Further, we introduce heuristics to consider the weight differences between the layers of the original model and the reconstructed layers from the codebooks to maximize the compression rate and accuracy.
Finally, we develop an efficient model execution and switching framework to operate multiple heterogeneous models targeted for different tasks, reducing the overhead of context switching (i.e., model swap between tasks) at run-time.


YONO is comprised of two components. The first component is an offline phase in which a shared PQ codebook is learned and multiple models are incorporated. We implement the offline phase of our system on a server. The second component is an online phase in which multiple heterogeneous models are deployed on an extremely resource-constrained device (MCUs). To evaluate YONO, we first evaluated four image datasets and one audio dataset used in state-of-the-art prior work on heterogeneous MTL~\cite{lee_fast_2020} for a fair comparison. We show that YONO achieves high accuracy of 93.7\% on average across the five datasets, which is a 15.4\% improvement over~\cite{lee_fast_2020} due to our usage of the optimized network architecture (see \S\ref{subsec:two modality} for detail) and is very close to the accuracy of the uncompressed models (0.4\% loss in accuracy). 
Further, to evaluate the scalability of YONO to other modalities, we include data from  modalities such as accelerometer signals from Inertial Movement Units (IMU) for human activity recognition (HAR) and surface electromyography (sEMG) signals for gesture recognition (GR). We then demonstrate that YONO effectively retains the accuracy of the uncompressed models across all the employed datasets of four different modalities (Image, Audio, IMU, sEMG).
Next, to evaluate the generalizability of the learned codebooks of YONO, we apply YONO to compress new models trained on unseen datasets during the codebook learning in the offline phase. Surprisingly, YONO can maintain the accuracy of the uncompressed models and achieve a 12.37$\times$ compression ratio (53.1\% higher than~\cite{lee_fast_2020}).
\yd{Finally, we evaluate the online component of YONO on the largest model and the smallest model to show the upper bound and lower bound results, respectively. We employ an MCU, STM32H747XI (see Section~\ref{sec:implementation} for details), and demonstrate that YONO enables an efficient in-memory execution (latency of 16-159 ms and energy consumption of 3.8-37.9 mJ per operation) and model loading/swap framework for task switching (showing reductions of 93.3-94.5\% in latency and 93.9-95.0\% in energy consumption compared to the method using external storage access).}

\section{YONO}\label{sec:overview}

In this section, we first present the overview of our multitasking system, YONO (\S\ref{subsec:overview}). Then, we introduce the background on PQ and its applications on  single model compression (\S\ref{subsec:pq}). We then explain how we utilize PQ to compress multiple heterogeneous networks into a pair of codebooks. The networks can be of any arbitrary architecture that consists of fully connected layers and convolutional layers. After that, we present our novel network optimization process to ensure the performance of the compressed networks remain close to original models (\S\ref{subsec:optimization}). On top of that, based on an observation (detailed in \S\ref{subsec:heuristics}), we further propose optimization heuristics to maximize the performance gain with a minimal loss of the compression rate when using PQ-based compression. Finally, we describe our in-memory execution and model swapping framework on MCUs (\S\ref{subsec:in_memory}).

\subsection{Overview}\label{subsec:overview}
In this subsection, we describe the overview of YONO that learns codebooks to represent the weights of multiple heterogeneous neural networks as well as enable on-chip memory operations on resource-constrained devices. In particular, YONO is composed of two components: (1) an offline phase where YONO learns a pair of codebooks on pretrained neural networks using PQ (will be explained in detail in \S\ref{subsec:pq}) and (2) an online phase where YONO enables on-chip execution such as model execution and model loading/swapping.
Note that we assume that the overall size of multiple neural networks is larger than the operational limit of the on-chip eFlash memory and SRAM of the targeted IoT devices. For example, in Section~\ref{sec:evaluation}, we employ seven different models with a total size of 3.84 MB and evaluate our framework on MCU (STM32H747XI), which strictly has only 512 KB of SRAM and 1 MB of eFlash.


\begin{figure*}[t]
  \centering
    \includegraphics[width=0.8\textwidth]{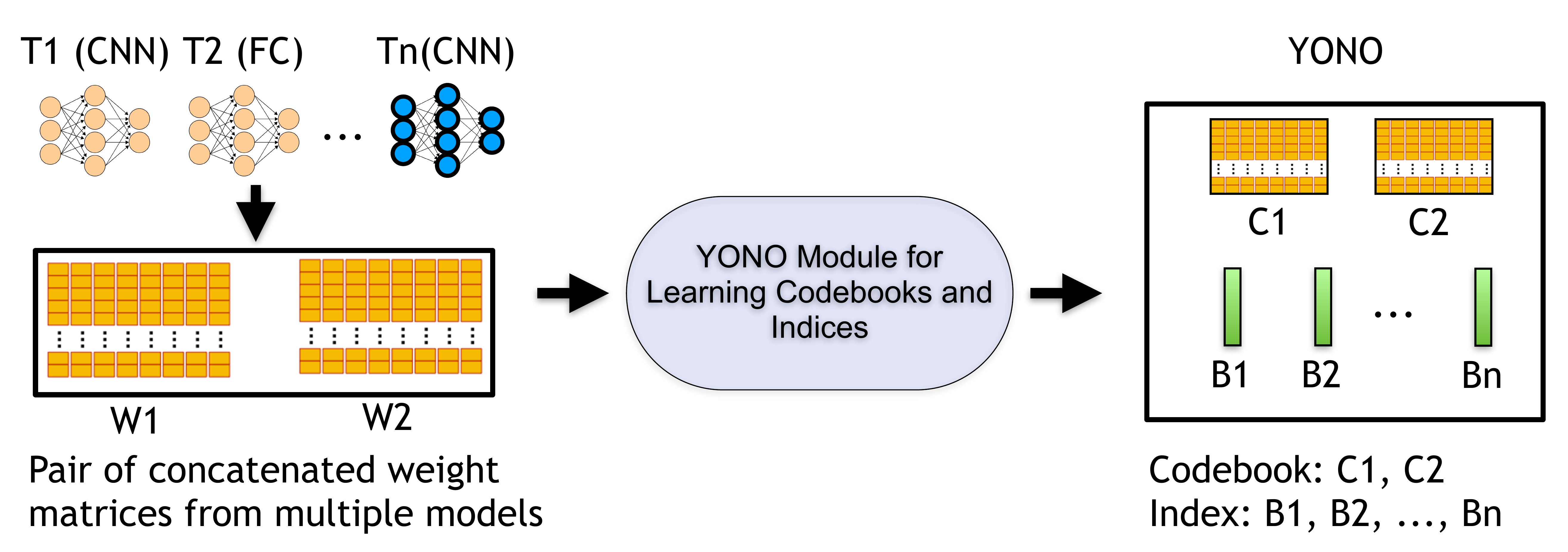}
  \caption{
  Overview of the offline component of YONO. The offline module employs PQ to learn a pair of codebooks and identify indices to represent multiple heterogeneous neural networks. This module incorporates our novel optimization process and heuristics to minimize the accuracy loss compared to the original models.
  }
  \label{fig:YONO_offline_module}
\end{figure*}



\subsection{Product Quantization and Compressing Single Neural Network}\label{subsec:pq}

We now provide an introduction to PQ and how it is used to compress a single model. 
PQ can be considered a special case of vector quantization (VQ)~\cite{gray_vector_1984}, in which it attempts to find the nearest codeword, $\mathbf{c}$, to encode a given vector, $\mathbf{w}$. 
\yd{Suppose we are given a codebook, $\mathbf{C}$, that contains a set of representative codewords, we can reconstruct/approximate the given vector $\mathbf{w}$ by using $\mathbf{c}$ and its associated index in the codebook.
Thus, given a vector $\mathbf{w} \in \mathbb{R}^d$ to be encoded, the encoding problem of VQ can be formulated as follows.}


\begin{equation}\label{eq:vq}
    \underset{b}{\mathrm{argmin}} \norm{ \mathbf{w} - \textbf{C}b }^2 
\end{equation}
where $\mathbf{C}$ is a $d$-by-$K$ matrix containing $K$ codewords of length $d$, and $b$ is called a code (i.e., index of codebook pointing to a codeword, $\mathbf{c}$, nearest to the given vector, $\mathbf{w}$). $\norm{\cdot}$ is a $l_2$ norm. \yd{Solving Equation~\ref{eq:vq} is equivalent to searching the nearest codeword. Besides, the codebook, $\mathbf{C}$, is learned by running the standard k-means clustering over all the given vectors~\cite{jegou_product_2011}.}

The PQ is a particular case of VQ when the learned codebook is the Cartesian product of sub-codebooks. Given that there are two sub-codebooks, the encoding problem of PQ is as follows.
\begin{equation}
\begin{aligned}
\label{eq:pq}
    \underset{b}{\mathrm{argmin}} \norm{ \mathbf{w} - \mathbf{C}b }^2, \\ 
    s.t. \quad \mathbf{C} = \mathbf{C}_1 \times \mathbf{C}_2
\end{aligned}
\end{equation}
where $\mathbf{C}_1$ and $\mathbf{C}_2$ are two sub-codebooks of $\frac{d}{2}$-by-$K$ matrices. Since any codeword of $\mathbf{C}$ is now the concatenation of a codeword of $\mathbf{C}_1$ and a codeword of $\mathbf{C}_2$, PQ can have $K^2$ different combinations of codewords. If a vector is divided into $M$ partitions, then PQ can have $K^M$ combinations of codewords. \yd{The number of sub-codebooks, $M$, can be any number between 1 and the length of the given vector, $d$ (e.g., 1, 2, …, $d$). When $M$ is set to 1, it is VQ. When $M$ is set to $d$, it is equivalent to the scalar k-means algorithm.}

We now describe how the encoding problem of PQ can be applied to compress a neural network. It is because instead of storing weight matrix $\mathbf{W}$ of any layer in neural networks explicitly, we can learn an encoding $\mathcal{B}(\mathbf{W})$ that needs much less storage space. Using the found encoding $\mathcal{B}$ and a learned codebook $\mathbf{C}$ based on PQ, we can reconstruct $\mathbf{\widehat{W}}$ which approximates the original weight matrix $\mathbf{W}$ of the layer. If we can find $\mathbf{\widehat{W}}$ close enough to $\mathbf{W}$, the reconstructed layer of a neural network will perform normally as demonstrated in prior works using PQ to compress a single neural network~\cite{gong_compressing_2014,stock_training_2020}.

\subsection{Compressing Multiple Heterogeneous Networks}\label{subsec:multiple_model}

As described in \S\ref{subsec:pq}, PQ is typically used to compress a single model in machine learning literature~\cite{stock_and_2019,martinez_permute_2021}. In prior works, each layer is replaced by one small-sized codebook (e.g., K=256, D=8, M=1), and a high compression rate and little performance loss are achieved in large computer vision models with more than 10 M parameters (e.g., ResNet50~\cite{he_deep_resnet}). \yd{However, in small-sized models that are specially designed to be used on MCUs (i.e., the number of parameters is at most around 500K-1M), the same approach (having a codebook for each layer) no longer provides a high compression rate due to the overhead of storing many codebooks.} 
Therefore, in our system, we propose to apply PQ to one or multiple neural networks while only sharing a pair of the learned codebooks to maximize the compression ratio. We will explain how we ensure high performance of the compressed models in the next subsections (\S\ref{subsec:optimization} and \S\ref{subsec:heuristics}). 

As in Figure~\ref{fig:YONO_offline_module}, we first concatenate weights of all the models of different tasks (i.e., $T_1, T_2, ..., T_n$). Then, we construct two weight matrices, $W_1$ and $W_2$, so that YONO takes into account spatial information of convolutional layer kernels as in other prior works~\cite{stock_training_2020}. \yd{For one weight matrix, $W_1$, we combine convolutional layers with a kernel size of $3\times3$. Then, in the other weight matrix, $W_2$, we concatenate convolutional layers with kernel size $1\times1$ and fully-connected layers. Then these concatenated weight matrices, $W_1$ and $W_2$,  are given as an input to learn codebooks, $C_1$ and $C_2$, for different kernel sizes, respectively.} 
\yd{Note that we also observed that neglecting such information in learning codebooks leads to worse performance.} In our system design, we select kernel sizes of $3\times3$ and $1\times1$ as those are widely used kernel sizes in many of the optimized network architectures~\cite{howard_mobilenets_2017,sandler_mobilenetv2_2018,ma_shufflenet_2018}. Also, since FC layers are essentially the same as point-wise convolution operation (i.e., kernel size of $1\times1$), we combine weights of FC layers together with those of $1\times1$ kernel convolution layers. 
\yd{Besides, we set M to 2 throughout our evaluation so that YONO can leverage the implicit codebook size of $K^M$. We observed that when M is 1, the codebook is not generalizable enough to compress multiple neural networks. When M is set to 3, the overhead of the codebooks decreases the compression rate without providing much accuracy benefit.}


\subsection{Network Optimization}\label{subsec:optimization}

After learning a pair of codebooks for multiple models as in \S\ref{subsec:multiple_model}, YONO performs finetuning on the reconstructed model in order to adjust the loss of information due to the compression (see Algorithm~\ref{alg:optimization}). As studied in~\cite{zhang_systematic_2018}, weights in the first and last layer of a model are the most important. Thus, in the finetuning stage, we select the first and last layer of a model and finetune them (Lines 2-4). The finetuning step largely recovers the accuracy of the original model by re-adjusting the first and last layer of the model according to the different weights induced by the codebooks. However, as we will show in our evaluation in Section~\ref{sec:evaluation} (this incurs 2-8\% accuracy loss), a simple extension of PQ to multiple heterogeneous neural networks with a finetuning step cannot ensure high accuracy due to the increased weight differences between original models' weight matrices $\mathbf{W}_{T_1,...,T_n}$ and reconstructed models' weight matrices $\mathbf{\widehat{W}}_{T_1,...,T_n}$ although it shows a high compression~rate.


Therefore, we introduce an optimization process to improve the performance of the decompressed models. As discussed in prior works~\cite{gong_compressing_2014,stock_and_2019}, in general, higher weight differences (i.e., errors) result in increased loss of accuracy. Thus, to minimize the impact of the weight differences, we adopt to use the iterative optimization procedure, inspired by the Expectation-Maximization (EM) algorithm~\cite{dempster_maximum_1977} and prior work~\cite{stock_and_2019}. We iteratively adjust the weight drifts by reassigning indices on the updated weights from finetuning as the E-step (Lines 12-13) and by finetuning several selected layers (e.g., first and last layers) as the M-step (Lines 14-17). 
\yd{Note that our optimization procedure is novel in that (i) we perform network optimization across multiple heterogeneous networks and (ii) we do not update codewords in our learned codebooks since we want our codebooks to be generalizable to compress unseen models and datasets during the codebook learning procedure, different from single model compression methods~\cite{gong_compressing_2014,stock_and_2019,stock_training_2020,martinez_permute_2021}.}
In Section~\ref{sec:evaluation}, we demonstrate the generalizability of our learned codebooks and our system on new models that are trained on new datasets that YONO did not see in its codebook learning.

\begin{algorithm}[!t]
\caption{YONO Network optimization and heuristics for a given task $t$}
\label{alg:optimization}
\SetAlgoNoLine
\DontPrintSemicolon
  \KwInput{Model weights $\mathbf{W}$, model indices $\mathbf{b}$, PQ codebooks $\mathbf{C}$, the number of layers $L$, error threshold $\epsilon$, heuristics}
  \KwOutput{Reconstructed model weights $\mathbf{\widehat{W}}$, model indices $\mathbf{\hat{b}}$}
  \KwData{Train data $\mathcal{D}^{TRAIN}$, Test data $\mathcal{D}^{TEST}$}
    
    \tcc{Perform an initial finetuning step}
    $\mathbf{\widehat{W}} \leftarrow \mathbf{C}(\mathbf{b})$ \tcp*{reconstruct a model via codebooks and indices}
    
    \SetAlgoVlined
    \For{$\ell = 2, ..., L-1$}
    {
        $\textnormal{FreezeWeights}(\mathbf{\widehat{W}}^{\ell})$\;
    }
    \tcp{run network training (e.g., BackProp) with loss function}
    $\textnormal{Finetune}(\mathbf{\widehat{W}}, \mathcal{D}^{TRAIN})$\; 
    
    $acc\_orig \leftarrow \textnormal{Evaluate}(\mathbf{W}, \mathcal{D}^{TEST}))$\;
    $acc\_recon \leftarrow \textnormal{Evaluate}(\mathbf{\widehat{W}}, \mathcal{D}^{TEST}))$\;
    
    \If{$acc\_orig - \epsilon \leq acc\_recon$}{
        \KwRet $\mathbf{\widehat{W}}, \mathbf{b}$\;
    }
    \tcc{Perform a further network optimization step}
    $S \leftarrow (1,L)$ \tcp{finetuning layer set}
    $\mathbf{\hat{b}} \leftarrow \mathbf{b}$\;
    \For{$i = 1, ..., L-2$}
    {
        \tcp{E-step: code re-assignment}
        \For{$\ell \notin S$}{
            $\mathbf{\hat{b}^{\ell}} \leftarrow \underset{b \in \mathbf{\hat{b}}^{\ell}}{\mathrm{argmin}} \norm{ \mathbf{\hat{w}}^{\ell} - \mathbf{C}b }^2$
        }
        \tcp{M-step: model update}
        $\mathbf{\widehat{W}} \leftarrow \mathbf{C}(\mathbf{\hat{b}})$\;
        \For{$\ell \notin S$}
        {
            $\textnormal{FreezeWeights}(\mathbf{\widehat{W}}^{\ell})$\;
        }
        $\textnormal{Finetune}(\mathbf{\widehat{W}}, \mathcal{D}^{TRAIN})$\; 
        $acc\_orig \leftarrow \textnormal{Evaluate}(\mathbf{W}, \mathcal{D}^{TEST}))$\;
        $acc\_recon \leftarrow \textnormal{Evaluate}(\mathbf{\widehat{W}}, \mathcal{D}^{TEST}))$\;
        \If{$acc\_orig - \epsilon \leq acc\_recon$}{
            \KwRet $\mathbf{\widehat{W}}, \mathbf{\hat{b}}$\;
        }
        \If{ \textnormal{heuristics is} \textit{OURS}} 
        {
            \tcp{choose a layer to finetune based on our heuristics}
            $\ell \leftarrow \underset{\ell}{\mathrm{argmax}} \norm{ \mathbf{W}^{\ell} - \mathbf{\widehat{W}}^{\ell} }^2 / N^\ell$\;
            $S \leftarrow (S,\ell)$
        }
    }
    
\end{algorithm}

\subsection{Optimization Heuristics}\label{subsec:heuristics}

In addition, we further propose an optimization heuristic that can maximize performance improvement while ensuring a high compression rate.
We observed that weight differences of each layer ($\mathbf{W}$ and $\mathbf{\widehat{W}}$) are not uniformly distributed. Besides, the number of parameters in each layer is considerably different. For example, MicroNet-KWS-M~\cite{banbury_micronets_2021} (we adopt this network architecture in our evaluation. Refer to Section~\ref{sec:evaluation} for detail) contains 12 convolutional and FC layers. Among them, one convolutional layer has a 4-dimensional weight matrix ($\mathbf{W} \in \mathbb{R}^{C_{cout}\times C_{in} \times k \times k }$) with a size of \{140, 1, 3, 3\} which has 1,260 parameters, whereas another convolutional layer in the same model can have weight matrix with a size of \{196, 112, 1, 1\} which has 21,952 parameters. The latter has 17.4 times more parameters than the former. 
Thus, based on this observation, we propose our novel optimization heuristic to select layers for finetuning that have the largest weight difference and contain the least number of parameters (refer to Lines 22-24 in Algorithm~\ref{alg:optimization}). 
Hence, given a network $\mathbf{W}$ with $L$ layers, we attempt to find a layer $\ell$ as follows. 
\begin{equation}
	\underset{\ell}{\mathrm{argmax}} \norm{ \mathbf{W}^{\ell} - \mathbf{\widehat{W}}^{\ell} }^2 / N^\ell
\end{equation}
where $\mathbf{W}^{\ell} - \mathbf{\widehat{W}}^{\ell}$ is a weight difference of weight matrices of the layer $\ell$, and $N^{\ell}$ is the number of the parameters of the layer $\ell$. 

\yd{In summary, through the optimization heuristics, YONO identifies a layer with the highest weight difference per parameter. After that, YONO finetunes the identified layer using our network optimization process introduced in \S\ref{subsec:optimization}. The process continues until the reconstructed model's accuracy is recovered to the target accuracy (Lines 20-21), i.e., accuracy loss is less than a given threshold $\epsilon$ (e.g., 2-3\% in our evaluation). The number of layers to be finetuned is less than or equal to three in most cases.
This process helps YONO maximize the compression ratio (small storage overhead) while retaining the accuracy of its compressed models close to their corresponding original (uncompressed) models.}
\yd{Note that the finetuned layers are then quantized into 8-bit integers in the online component of YONO as described in the next subsection.}


\subsection{In-memory Execution and Model Swap Framework on MCUs}\label{subsec:in_memory}
\begin{figure}[t]
  \centering
    \includegraphics[width=0.3\textwidth]{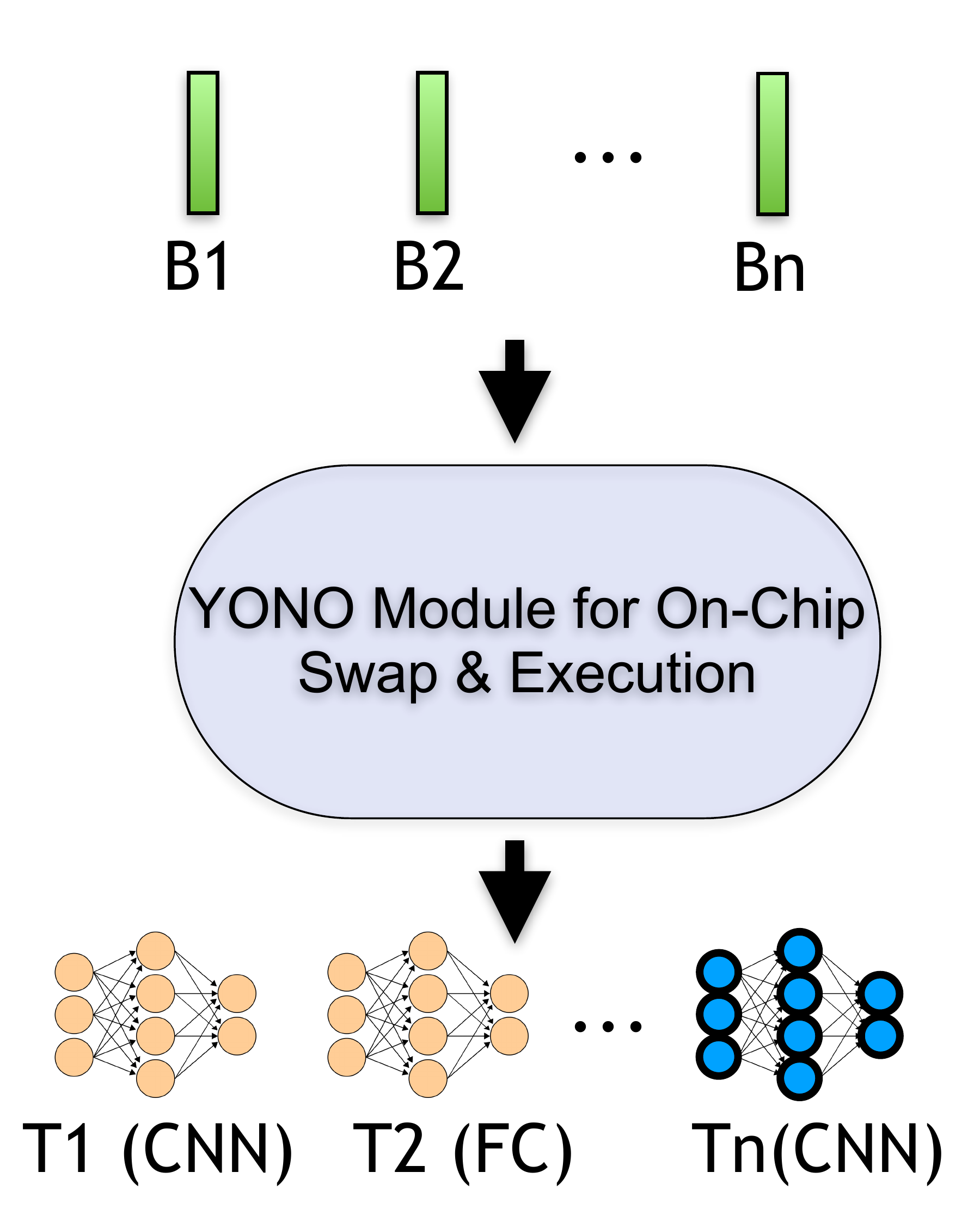}
  \caption{
  Overview of the online component of YONO. The online module enables fast and efficient model loading/swap and in-memory execution.
  }
  \label{fig:YONO_online_module}
\end{figure}

Having established the offline component of YONO, we now turn our attention to the online component of our system. At runtime, the online component of YONO enables the fast and efficient in-memory execution and model swap of multiple heterogeneous neural networks. Figure~\ref{fig:YONO_online_module} illustrates the overview of the online component of YONO.

\textbf{Data Structure for Deployment on MCUs:} To begin with, we describe the data structures that are necessary for deploying ML models on MCUs. First, YONO requires one pair of learned PQ codebooks, model indices, and other relevant information to reconstruct a model. In addition, YONO needs a task executor to run the reconstructed model in-memory and a task switcher to swap an in-memory model to another reconstructed model.

\textbf{Learned Codebooks:} As described in subsections \S\ref{subsec:pq}-\ref{subsec:heuristics}, YONO learns a pair of codebooks by applying PQ on multiple heterogeneous neural networks with our novel optimization procedure. Since  SRAM is a scarce resource on MCUs, the codebooks are stored on eFlash. Also, because the codebooks are shared across different models compressed by YONO and static during runtime, they are stored on the read-only memory of eFlash.

\textbf{Model Indices and Other Elements:} Once a model is compressed through our system, YONO generates model indices that correspond to the weights of an original model via the learned codebooks and other relevant elements necessary to reconstruct the uncompressed model. For example, relevant elements include model architecture, operators, quantization information, and so on.

\textbf{Task Executor:} We now present the explanation of our task executor. As we adopt TensorFlow Lite for Microcontrollers (TFLM)~\cite{robert_tflm_mlsys2021} to run the deployed model on MCUs, YONO also follows its model representation and interpreter-based task execution. As model representation on MCUs, the stored schema of data and values represent the model. The schema is designed for storage efficiency and fast access on mobile and embedded platforms. Therefore, it has some features that help ease the development of MCUs. For example, operations are in a topologically sorted list instead of a directed-acyclic graph, making conducting calculations be a simple looping through the operation list in order. In addition, YONO adopts interpreter-based task execution by relying on TFLM. Thus, the interpreter refers to the schema of the model representation and loads a model. After that, the interpreter handles operations to execute. Since YONO adopts an interpreter-based task executor and loads a model in the main memory for execution, YONO allows model switching at run time, which is not allowed with the code-generator-based compiler method~\cite{lin_mcunet_2020} because this method requires recompilation to switch a model.

\textbf{Task Switcher:} When a task needs to be switched (e.g., the target application is switched from image classification to voice command recognition), YONO replaces the loaded model in the memory with a new model to be executed. Using the same memory space between previous and new models, YONO can operate multiple models within a limited memory budget of SRAM. In addition, since YONO does on-chip memory operations to perform execution and model swap, YONO improves the response time and end-to-end execution time of different applications. It is because the access time to secondary storage devices is slower than that to internal memory and primary storage. Moreover, a system relying on external storage devices may have unpredictable overheads. For example, disk-writes on storage devices like flash and solid-state drives need to erase an entire block before a write operation.

\textbf{Model Reconstruction:} We now describe our model reconstruction scheme. To reconstruct a model, YONO utilizes the PQ codebooks, indices, and relevant elements, such as batch normalization layer's mean and variance, quantization information, stored in eFlash. The overall process is as follows. First, YONO retrieves model weights by matching indices of a model to be loaded on the main memory and its corresponding codewords of the PQ codebooks. 
Secondly, YONO loads relevant elements of the model and then writes this information and model weights to the preallocated memory address for the model on the main memory.

\yd{In addition, each value of the learned codewords in the PQ codebooks is stored in 16-bit float instead of 32-bit float type to further reduce the storage requirements on eFlash. In contrast, the weights of the model loaded on the main memory and executed need to be quantized to 8-bit integers. Thus, while loading each layer of the model, YONO converts 16-bit floats to 8-bit integers using the saved quantization information. Specifically, we use the quantization scheme used in~\cite{jacob_quantization_2018} to minimize the information loss in quantization. We utilize an affine mapping of integer q to real number r for constant quantization parameters $S$ and $Z$, i.e.,
$r = S (q - Z)$. $S$ denotes the scale of an arbitrary positive real number. $Z$ denotes zero-point of the same type as quantized value q, corresponding to the real value 0. As a result, the reconstructed model in the online component is based on 8-bit integers, and thus the use of codebooks does not affect computations of model execution.}

\section{System Implementation}\label{sec:implementation}
We introduce the hardware and software implementation of YONO. 

\textbf{Hardware.} The offline component of our system is implemented and tested on a Linux server equipped with an Intel Xeon Gold 5218 CPU and NVIDIA Quadro RTX 8000 GPU. \yd{This component is used to learn PQ codebooks and find indices for each model to be compressed. Then, the online component of our system is implemented and evaluated on an MCU, STM32H747XI, having two cores (ARM Cortex M4 and M7) with 1 MB SRAM and 2 MB eFlash in total.} However, our implementation of YONO uses only one core (ARM Cortex M7) since MCUs are typically equipped with one CPU core. We restrict the usage space of SRAM and eFlash to 512 KB and 1 MB, respectively, to enforce stricter resource constraints. 





\textbf{Software.} We use PyTorch 1.6 (deep learning framework) and Faiss (PQ framework) to develop and evaluate the offline component of YONO on the Linux server. At the offline phase, we develop YONO using Python on the server and examine the accuracy of the models. In addition, we develop the online component of YONO using C++ on STM32H7 series MCUs. For running neural networks on MCUs, we rely on TFLM. Since eFlash memory of MCUs is read-only during runtime, YONO loads the model weights on SRAM (read-write during runtime) and swaps the models by replacing the models' weights using PQ codebooks and indices stored on eFlash. 
The binary size of our implementation on an MCU is only 0.41 MB, and the total size of PQ codebooks, indices, and other information to compress the eight heterogeneous networks evaluated in \S\ref{subsec:unseen data} is 0.35 MB. Note that the memory requirement of the seven models is 4.19 MB, which is 12.05$\times$ of what YONO requires and 4.19$\times$ of what typical MCUs with 1 MB storage can support.



\section{Evaluation}\label{sec:evaluation}

We now present the results of the evaluation on our system. \S\ref{subsec:experimental setup} describes our experimental setup. We evaluate the effectiveness of our system in the offline phase regarding the performance (i.e., accuracy) and compression rate of the compressed models in an MTL scenario. 
To make a comparison with prior work~\cite{lee_fast_2020} that tackles MTL of different neural networks, we begin with evaluating our system with the same datasets used in~\cite{lee_fast_2020} consisting of five datasets for two modalities (i.e., image and audio) (\S\ref{subsec:two modality}).
After that, we evaluate our system to what extent it can address multiple heterogeneous networks trained with different modalities. Thus, we employ four different modalities of data ((1) Image, (2) Audio, (3) IMU, (4) sEMG) by adding two more datasets in order to demonstrate the scalability of YONO on diverse modalities in \S\ref{subsec:four modality}. Further, to demonstrate the generalizabilty of YONO's learned codebooks, we select two additional datasets in each of the four modalities and evaluate our system to compress new models trained on these datasets that YONO did not learn during its codebook learning stage (\S\ref{subsec:unseen data}). Finally, we present the results of our online in-memory model execution and swap operations in \S\ref{subsec:online eval}.

\subsection{Experimental Setup}\label{subsec:experimental setup}

\subsubsection{Task}
Our target application scenarios are based on dealing with dissimilar multitask learning. For example, those applications are image classification, keyword spotting, human activity recognition, and gesture recognition.

\subsubsection{Evaluation Protocol} 
Following prior works~\cite{emotionsense, su_environment_2019}, 10\% of data is used as the test set and the remaining as the training set. In addition, to evaluate the effectiveness of the offline phase component of our system, we report the accuracy and compression rate of the compressed models using our system. We also use  compressed model's error rate (i.e., accuracy loss) compared to the original model. Then, to evaluate the efficiency of the online phase component of our system, we report the execution time and load/swap time of the models on MCU.

\subsubsection{Baseline Systems} 
To evaluate the effectiveness of our work, \textbf{YONO}, we include various baselines in our experiments as follows. 

\textbf{NWV:} Neural Weight Virtualization (NWV)~\cite{lee_fast_2020} is the state-of-the-art heterogeneous MTL system that treats weights of neural networks as consecutive memory locations which can be virtualized and shared by multiple models. Note that we use reported results of \cite{lee_fast_2020} on an MCU, which relies on simplified LeNet architecture.

\textbf{Scalar Quantization (Int8):}
\yd{This baseline compresses a single model by quantizing 32-bit floats into low-precision fixed-point representation (e.g., 8-bit)~\cite{jacob_quantization_2018,krishnamoorthi_quantizing_2018}. As in~\cite{krishnamoorthi_quantizing_2018}, we employ both post-training quantization and quantization-aware training schemes. We then report the results of the best-performing scheme in our evaluation. Besides, we only include 8-bit quantization as sub-byte datatypes (e.g., 4-bit or 2-bit) are not natively supported by MCUs~\cite{banbury_micronets_2021}. We leave sub-byte quantization as future work.}

\textbf{PQ-S:} This baseline uses PQ to compress a single model to a pair of the shared codebooks across layers in the model. As this baseline does not share the codebooks across multiple models, this can serve \yd{as a baseline for the single model compression and as the lower bound in compression ratio among the PQ variants}.

\textbf{PQ-M:} This baseline uses PQ to compress multiple heterogeneous models to a pair of the shared codebooks but does not apply our optimization process and heuristics as described in Section~\ref{sec:overview}. We include this to conduct an ablation study to evaluate the impact of the proposed optimization in our system. 

\textbf{PQ-MOpt:} This baseline uses PQ to compress multiple heterogeneous models to a pair of the shared codebooks and also apply the optimization process without the heuristics described in Section~\ref{sec:overview}. We include this to conduct an ablation study to evaluate the impact of the heuristics in our system.

\textbf{Uncompressed (Original):} An original model before compression. It is pretrained with available training data and serves as the upper bound in terms of the accuracy metric.


\subsection{Performance}\label{subsec:two modality}
Following~\cite{lee_fast_2020}, we start by evaluating YONO in MTL scenarios on two modalities: images and audio signals which are widely used data modalities in mobile sensing applications.


\begin{table}[t]
  \centering
  \caption{
  Summary of datasets, model architectures, mobile applications used in \S\ref{subsec:two modality} and \S\ref{subsec:four modality}.
  }
  \label{tab:data_model_arch_1}
  \resizebox{\columnwidth}{!}{%
  \begin{tabular}{ c  c c c }
    \toprule 
    \textbf{Modality} & \textbf{Dataset} & \textbf{Architecture} & \textbf{Mobile Application} \\
        \cmidrule(l){0-3}
    \multirow{4}{*}{Image} & MNIST    & LeNet          & Digit recognition \\
     & CIFAR-10 & MicroNet-AD    & Object recognition \\
     & SVHN     & MicroNet-AD    & Digit recognition \\
     & GTSRB    & MicroNet-AD    & Road sign recognition \\
        \cmidrule(l){0-3}
    Audio & GSC      & MicroNet-KWS & Keyword spotting \\
        \cmidrule(l){0-3}
    IMU  & HHAR      & MicroNet-AD & Activity recognition \\
        \cmidrule(l){0-3}
    sEMG & Ninapro DB2 & Lightweight CNN & Gesture recognition \\
    \bottomrule
  \end{tabular}
  }
  \vspace{-0.5cm}
\end{table}

\textbf{Datasets.} \yd{We employ the same datasets used in the prior work~\cite{lee_fast_2020} to make a fair comparison. First, four image datasets are employed, namely MNIST~\cite{mnist}, CIFAR-10~\cite{cifar10}, SVHN~\cite{svhn}, and GTSRB~\cite{gtsrb} associated with classifying objects of handwritten digits (grayscale), generic objects, numbers (RGB), and road signs, respectively. Then, one audio dataset of Google Speech Commands V2 (GSC)~\cite{gscv2} for keyword spotting is used.}

\textbf{Model Architecture.}  
We adopt optimized neural network architectures, designed to be used in the resource-constrained setting, such as variants of MicroNet~\cite{banbury_micronets_2021}, simplified LeNet used in~\cite{lee_fast_2020}. 
For MNIST, we use the simplified LeNet as it is used in~\cite{lee_fast_2020} and the accuracy of such LeNet variant is very high at ~98\%.  
For other datasets (CIFAR-10, SVHN, GTSRB, GSC), we use variants of MicroNet architecture to construct pretrained models. To identify a high-performing and yet lightweight model to operate on embedded and mobile devices, we conduct a hyper-parameter search based on different variants of MicroNet (e.g., small, medium, large models), lightweight convolutional neural network (CNN) architectures~\cite{kwon21_interspeech}, the number of convolutional filters. A basic convolutional layer consists of $3 \times 3$ convolution, batch normalization, and Rectified Linear Unit (ReLU). 
Then, as our final model architectures, we use MicroNet-KWS-M for GSC and MicroNet-AD-M (with the reduced number of convolutional filters \{192\}) for CIFAR-10, SVHN, GTSRB. Throughout model training for all of the datasets, ADAM optimizer~\cite{kingma_adam_2017} and learning rate of 0.001 are used. The datasets, architectures, and applications are summarized in Table~\ref{tab:data_model_arch_1}.

\begin{figure}[t]
  \centering
    \includegraphics[width=0.48\textwidth]{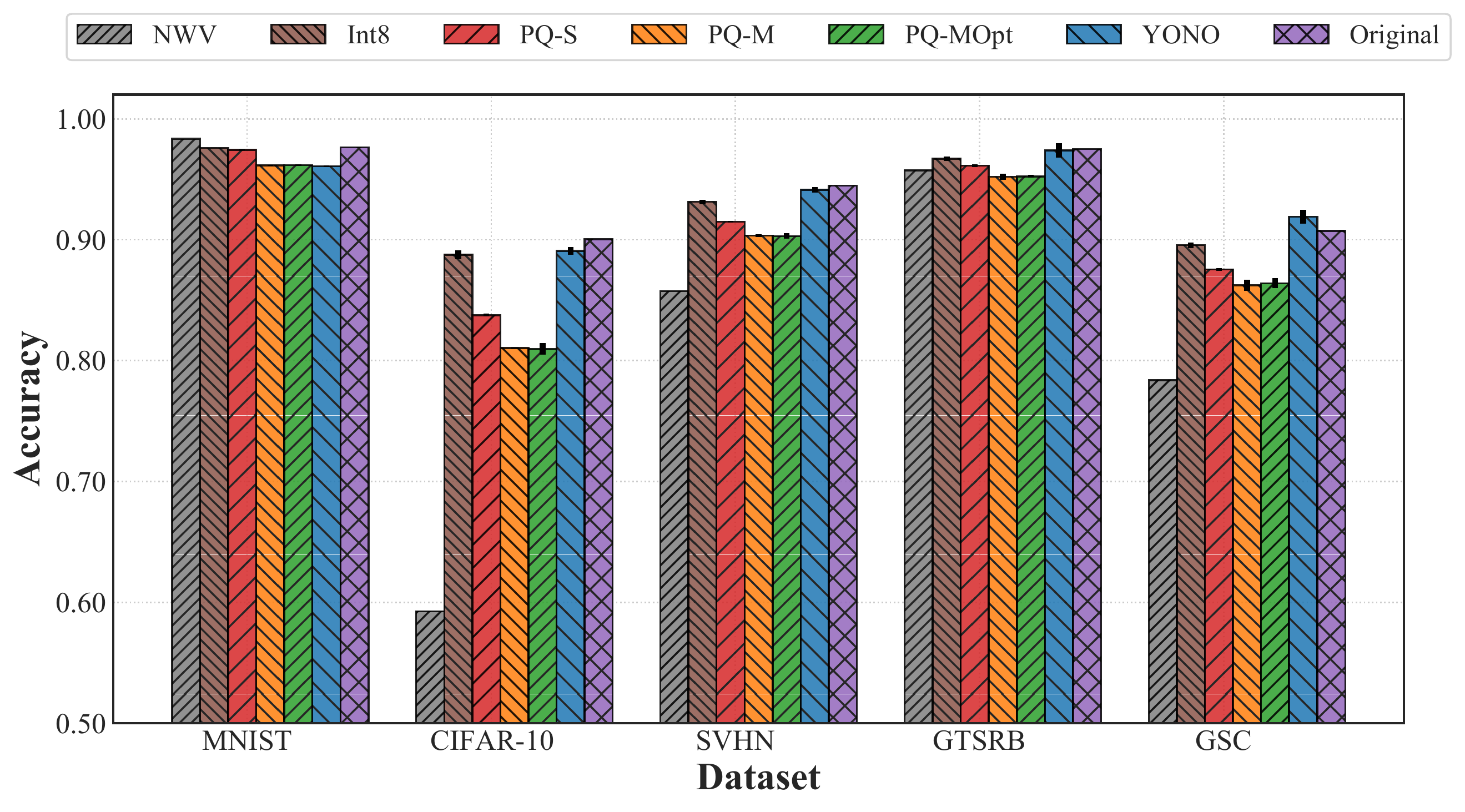}
  \caption{
  The inference accuracy of the heterogeneous MTL systems trained with five datasets of two modalities. Reported results are averaged over five trials, and standard-deviation intervals are depicted.
  }
  \label{fig:2modal_acc}
\end{figure}

\textbf{Accuracy.}
We show the accuracy results here. Figure~\ref{fig:2modal_acc} shows the accuracy of each baseline so that we can analyze the impact of our proposed techniques in our system. 
To begin with, the uncompressed (original) model serves as a performance upper bound. \yd{8-bit quantization and PQ-S achieve high accuracy close to that of the original model, showing a small average error rate of 0.9\% and 2.8\%, respectively, between each of the five models after compression and their corresponding original models.} However, in the case of the CIFAR-10 dataset, PQ-S shows high error rates of 6.3\% on average. This result indicates that the specialized codebooks which target only one model can help retain the performance of the original model in general but sometimes fail to retain it, as shown in the case of CIFAR-10. Besides, PQ-M shows an accuracy loss of 4.3\% on average. For CIFAR-10, it shows a high error rate of 9.0\%. In addition, although our proposed EM-based iterative network optimization procedure can help in improving the accuracy, PQ-MOpt still shows a substantial accuracy drop of 4.3\% on average. This result indicates that compressing multiple neural networks based on only one pair of codebooks is very challenging. However, YONO shows that its accuracy drop is minimal (i.e., an average error rate of 0.4\%). Interestingly, in the case of GSC, YONO outperforms the accuracy of the original model by 1.2\% where YONO benefits from sharing weights via PQ codebooks.


This result indicates that YONO can effectively retain the accuracy of original models as observed in the prior work on multiple MTL systems~\cite{lee_fast_2020} and other techniques focusing on a single model compression~\cite{jacob_quantization_2018,han_deep_2016,stock_and_2019}. Further, it is an interesting result because YONO can retain the accuracy of multiple heterogeneous models, which is more challenging given that simply performing MTL would lead to the accuracy drop as shown in the prior work, NWV~\cite{lee_fast_2020}. Also, note that differently from~\cite{lee_fast_2020} which used LeNet, we used optimized network architectures such as MicroNet and lightweight CNN that can execute on resource-constrained MCUs (refer to \S\ref{subsec:online eval}) and obtain very high accuracy. To name a few, the pretrained models in our work achieve 90.05\%, 94.48\%, 90.74\% on CIFAR-10, SVHN, GSC compared to 59.26\%, 85.74\%, 78.38\% reported in~\cite{lee_fast_2020}, respectively.


\begin{table}[t]
  \centering
  \caption{
  The compression efficiency of the heterogeneous MTL systems trained with five datasets of two modalities.
  }
  \label{tab:2_model_compress}
  \resizebox{1.01\columnwidth}{!}{%
  \begin{tabular}{ c c c c c c c c }
    \toprule 
     & \textbf{NWV}~\cite{lee_fast_2020} & \textbf{Int8} & \textbf{PQ-S}  & \textbf{PQ-M} & \textbf{PQ-MOpt} & \textbf{YONO} & \textbf{Original}\\
        \cmidrule(l){0-7}
    Ratio & 8.08$\times$ & 3.04$\times$ & 9.47$\times$ & 12.07$\times$ & 12.07$\times$ & 11.57$\times$ & 1$\times$\\
    Size  & 0.13 MB & 0.96 MB & 0.31 MB & 0.24 MB & 0.24 MB & 0.25 MB & 2.91 MB \\
    \bottomrule
  \end{tabular}
  }
\end{table}

\textbf{Compression Efficiency.} Table~\ref{tab:2_model_compress} shows the overall efficiency in compressing heterogeneous networks trained with five datasets of two modalities.
First, the combined storage overhead of the five uncompressed models is 2.91 MB which is three times the capacity of our target MCU's storage, which is 1 MB at maximum. However, considering that to perform an inference on MCUs, it is required to have a space for program codes of TFLM, input and output peripherals, input and output buffers, and other variables, etc., the space used to store models needs to be below the storage size of 1 MB. Thus, it is impossible to put those five models on an MCU and run multitask applications using the uncompressed models.
\yd{8-bit quantization shows the lowest compression rate of 3.04$\times$ among all the evaluated methods, and its storage size (0.96 MB) is just below the limit of our employed MCU.}
Then, PQ-S shows a moderate compression rate of 9.47$\times$ and decreases the required storage size down to 0.31 MB and thus can reside on an MCU. Other baseline systems, PQ-M and PQ-MOpt, show a high compression rate of 12.07$\times$ and reduce the storage requirement to 0.24 MB. This is because PQ-M and PQ-MOpt share the same codebooks across the different applications. However, the savings in storage come at the expense of loss of accuracy, as seen in the accuracy results discussed before. In contrast, YONO achieves the best of both worlds, demonstrating a high compression rate close to PQ-M and PQ-MOpt and negligible accuracy loss compared to the uncompressed models. YONO obtains a 11.57$\times$ compression rate and decreases the storage overhead to 0.25 MB, showing a higher compression rate than NWV~\cite{lee_fast_2020}. 

\textit{Overall, the results indicate that YONO can enable running multi-task applications on MCUs while retaining high accuracy and low storage footprints.}



\subsection{Scalability}\label{subsec:four modality}
In this subsection, we apply YONO on seven datasets consisting of four different data modalities ((1) Image, (2) Audio, (3) IMU, (4) sEMG) to investigate to what extent our system can effectively compress multiple networks trained on different data modalities without losing its accuracy and compressive power. \yd{We select IMU and sEMG as additional modalities because they are also widely used in mobile sensing applications~\cite{hhar, fan_what_2018}.}


\textbf{Datasets.} \yd{On top of the five datasets used in the previous subsection, we add two datasets of two additional modalities: HHAR~\cite{hhar} and Ninapro DB2~\cite{ninadb2_ninadb3}, corresponding to activity recognition (based on IMU) and gesture recognition (based on sEMG), respectively. The HHAR and Ninapro DB2 datasets are some of the most widely used HAR and sEMG datasets, respectively. }

\textbf{Model Architecture.}
To identify the right model architecture for each dataset, we adopt to use the optimized neural architectures and also conduct a hyper-parameter search as described in \S\ref{subsec:two modality}. Then, we select the model which shows the best performance. As a result, we use MicroNet-AD for HHAR and lightweight CNN architecture for Ninapro DB2 (see Table~\ref{tab:data_model_arch_1} for detail).


\begin{figure}[t]
  \centering
    \includegraphics[width=0.48\textwidth]{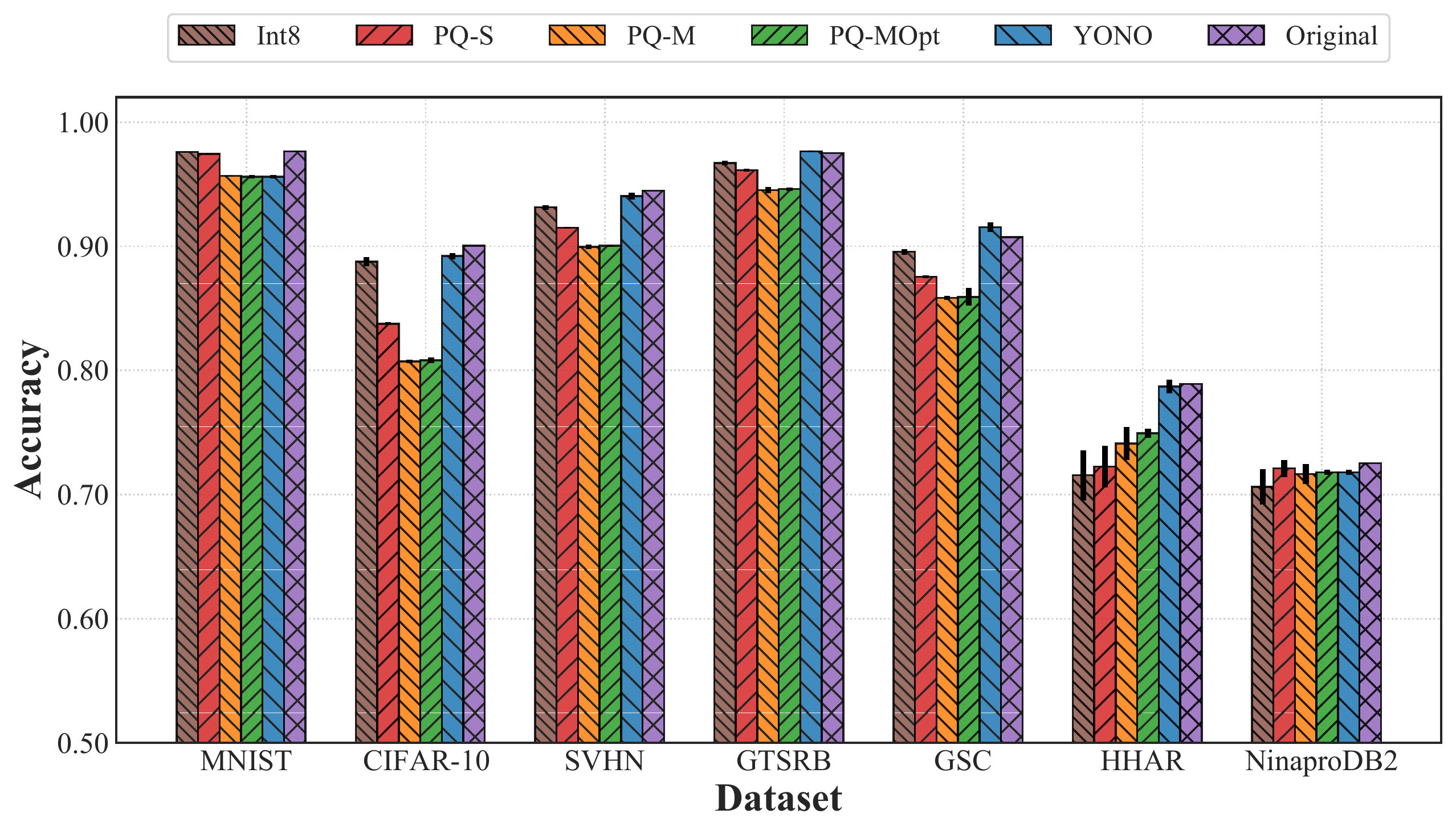}
  \caption{
  The inference accuracy of the heterogeneous MTL systems trained with seven datasets of four modalities. Reported results are averaged over five trials, and standard-deviation intervals are depicted.
  }
  \label{fig:4modal_acc}
\end{figure}

\textbf{Accuracy.}
Figure~\ref{fig:4modal_acc} presents the accuracy results of the seven datasets of four modalities so that we examine the scalability of YONO to various modalities of data. 
Overall, the accuracy of reconstructed models from baseline systems and YONO is slightly improved since the error rates on the new datasets are smaller than those of the other five datasets. Also, accuracy results of baseline systems and YONO reach similar observations as in \S\ref{subsec:two modality}. 
\yd{8-bit quantization shows a small average error rate of 2.0\% but a relatively high accuracy variance for HHAR.}
Also, PQ-S achieves high accuracy close to that of the uncompressed models with small error rates of 3.0\% on average, whereas it shows high error in CIFAR-10. 
Then, the PQ-M and PQ-MOpt systems present an average error rate of 4.2\% and 4.0\% respectively, indicating that our proposed EM-based iterative network optimization procedure help improve the accuracy but still falls short of achieving the original model's accuracy. Also, in this setting, YONO performs the best and shows an negligible accuracy loss of 0.5\% on average. 

\begin{table}[t]
  \centering
  \caption{
  The compression efficiency of the heterogeneous MTL systems trained with seven datasets of four modalities.
  }
  \label{tab:4_model_compress}
  \resizebox{\columnwidth}{!}{%
  \begin{tabular}{ c c c c c c c }
    \toprule 
     & \textbf{Int8} & \textbf{PQ-S}  & \textbf{PQ-M} & \textbf{PQ-MOpt} & \textbf{YONO} &  \textbf{Original} \\
        \cmidrule(l){0-6}
    Ratio & 2.96$\times$ & 9.27$\times$ & 12.29$\times$ & 12.29$\times$ & 11.77$\times$ & 1$\times$ \\
    Size  & 1.27 MB & 0.41 MB & 0.31 MB & 0.31 MB & 0.32 MB & 3.76 MB \\
    \bottomrule
  \end{tabular}
  }
\end{table}

\textbf{Compression Efficiency.} Table~\ref{tab:4_model_compress} shows the overall efficiency in compressing heterogeneous models trained with seven datasets of four modalities. Similar to the compression results in \S\ref{subsec:two modality}, the total size of the seven uncompressed models (3.76 MB) is larger than the storage budget for our target MCU. 
\yd{In the case of 8-bit quantization, the required storage size of the seven compressed models is 1.27 MB, larger than our storage budget of 1 MB. This result indicates that 8-bit quantization is not suitable for operating many heterogeneous neural networks simultaneously on our target MCU.}
However, YONO requires at most 0.32 MB. Since our system can effectively compress multiple heterogeneous models (showing 11.77$\times$ compression ratio), the incurred storage requirement is minimal. For example, when two additional models (for HHAR and Ninapro DB2) are included in an MTL system, YONO incurs only 0.07 MB additional overhead, whereas the original models' storage size increases by 0.85 MB.

\textit{To summarize, our results show that YONO is scalable as it can accommodate many applications utilizing  different input modalities while achieving high performance and small storage overhead.}

\subsection{Generalizability}\label{subsec:unseen data}
We now investigate the generalizability of our multitasking system on new models/datasets and different network architectures unseen during the codebook learning phase of the offline component. Specifically, we evaluate whether YONO can achieve high accuracy on the unseen models from new datasets using the same codebooks that are learned previously (\S\ref{subsec:four modality}). This can be particularly useful since the learned codebooks of YONO can still be utilized to compress unseen models in different network architectures from new datasets without learning new codebooks again whenever a user wants to incorporate a new task/dataset into the system. 
Also, note that since the codebooks are not modified, the reported results in \S\ref{subsec:four modality} are not affected, ensuring high accuracy on previous datasets.
Then, in \S\ref{subsec:unseen data}, we select two new datasets in each of the four modalities for a robust evaluation.


\textbf{Datasets.} In total, we add eight new datasets: two image datasets (1) FashionMNIST~\cite{fashion_mnist}, (2) STL-10~\cite{stl10}, and two audio datasets (3) EmotionSense~\cite{emotionsense}, (4) UrbanSound~\cite{urbansound}, and two HAR datasets (5) PAMAP2~\cite{pamap2}, (6) Skoda~\cite{skoda}, and lastly two sEMG datasets (7) Ninapro DB3~\cite{ninadb2_ninadb3} and (8) Ninapro DB6~\cite{ninadb6}. \yd{These are widely used real-world application datasets corresponding to classification problem as follows: (1) ten fashion items, (2) ten generic objects, (3) five emotions, (4) ten environmental sounds, (5) 12 activities, (6) ten activities, (7) ten gestures of amputees, (8) seven gestures of ordinary people, respectively.}

\textbf{Model Architecture.} To demonstrate that YONO can effectively address new network architectures that were not shown during the offline codebook learning phase, we include another widely used architecture, DS-CNN~\cite{zhang_hello_2017}, in our work. Then, we follow the same hyper-parameter search process as described in \S\ref{subsec:two modality}. Table~\ref{tab:data_model_arch_2} summarizes the identified network architectures for each dataset and its associated mobile application.

\begin{table}[t]
  \centering
  \caption{
  Summary of datasets, model architectures, mobile applications used in \S\ref{subsec:unseen data}.
  }
  \label{tab:data_model_arch_2}
  \resizebox{\columnwidth}{!}{%
  \begin{tabular}{ c  c c c }
    \toprule 
    \textbf{Modality} & \textbf{Dataset} & \textbf{Architecture} & \textbf{Mobile Application} \\
        \cmidrule(l){0-3}
    \multirow{2}{*}{Image} & FashionMNIST    & DS-CNN          & Object recognition \\
     & STL-10 & DS-CNN    & Object recognition \\
     \cmidrule(l){0-3}
    \multirow{2}{*}{Audio} & EmotionSense    & Lightweight CNN          & Emotion recognition \\
     & UrbanSound &DS-CNN    & Sound classification \\
     \cmidrule(l){0-3}
    \multirow{2}{*}{IMU} & PAMAP2    & MicroNet-AD          & Activity recognition \\
     & Skoda & MicroNet-AD    & Activity recognition \\
     \cmidrule(l){0-3}
    \multirow{2}{*}{sEMG} & Ninapro DB3    & Lightweight CNN          & Gesture recognition \\
     & Ninapro DB6 & MicroNet-AD    & Gesture recognition \\
    \bottomrule
  \end{tabular}
  }
\end{table}


\textbf{Accuracy.} 
Note that we exclude PQ-S as it needs to learn PQ codebooks on a given dataset and then perform network finetuning on the given dataset. However, in this scenario, the system needs to adapt to new (unseen) datasets. 
This point makes the scenario particularly challenging since an MTL system needs to incorporate unseen datasets and network architectures. Nonetheless, an MTL system that can address this challenge could become very useful in practice since it is adaptable.

\begin{figure}[t]
  \centering
    \includegraphics[width=0.48\textwidth]{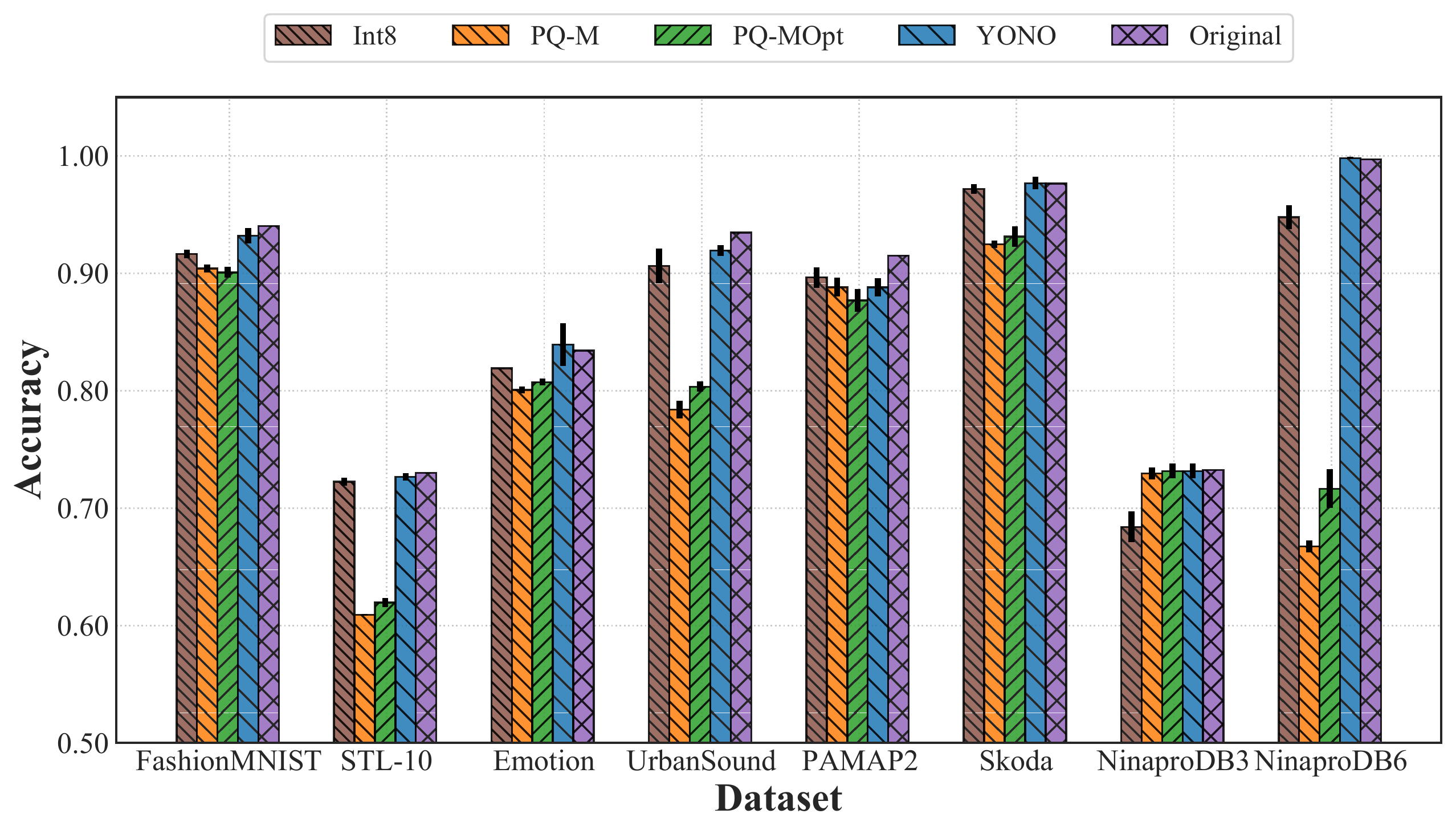}
  \caption{
  The inference accuracy of the heterogeneous MTL systems applied to unseen datasets of four modalities. Reported results are averaged over five trials, and standard-deviation intervals are depicted.
  }
  \label{fig:unseen_acc}
\end{figure}

To begin with, Figure~\ref{fig:unseen_acc} shows the accuracy results on the eight unseen datasets with diverse network architectures. 
\yd{8-bit quantization presents a moderate error rate of 2.5\% similar to the results in \S\ref{subsec:two modality} and \S\ref{subsec:four modality} as the current evaluation setup does not make a difference for the single model compression approach.}
Conversely, PQ-M shows a substantial accuracy drop (9.4\%) compared to the original model, which is worse than the previous two scenarios where it obtained error rates of 4.3\% and 4.2\%. In fact, on one dataset (Ninapro DB6), PQ-M shows a 33.0\% error rate. Although PQ-MOpt improves upon PQ-M, the amount of improvement is small. PQ-MOpt shows a 8.4\% accuracy drop on average compared to the original model. Also, for Ninapro DB6, the accuracy of PQ-MOpt shows a sharp decrease of 28.1\% compared to the original model, demonstrating the difficulty of this scenario.
Surprisingly, however, YONO does not experience a considerable accuracy loss. It shows only 0.6\% accuracy loss on average. 
Besides, YONO shows a low variance of accuracy loss across the employed datasets.
In fact, YONO even improves upon the accuracy of uncompressed models for some datasets such as EmotionSense, Skoda, and Ninapro DB6. These results highlights that YONO is capable of retaining the accuracy of original models even in the most challenging scenario of incorporating unseen datasets and architectures.


\begin{table}[t]
  \centering
  \caption{
  The compression efficiency of the heterogeneous MTL systems applied to unseen datasets of four modalities.
  }
  \label{tab:unseen_model_compress}
  \resizebox{0.9\columnwidth}{!}{%
  \begin{tabular}{ c c c c c c c }
    \toprule 
     & \textbf{Int8} & \textbf{PQ-M} & \textbf{PQ-MOpt} & \textbf{YONO} &  \textbf{Original} \\
        \cmidrule(l){0-5}
    Ratio & 2.80$\times$ & 13.60$\times$ & 13.60$\times$ & 12.37$\times$  & 1$\times$ \\
    Size  & 1.47 MB & 0.30 MB & 0.30 MB & 0.33 MB & 4.11 MB \\
    \bottomrule
  \end{tabular}
  }
\end{table}

\textbf{Compression Efficiency.} The compression results for heterogeneous models with eight unseen datasets are shown in Table~\ref{tab:unseen_model_compress}. The size of the uncompressed models is the largest, 4.11 MB, in this setup compared to \S\ref{subsec:two modality} and \S\ref{subsec:four modality}. YONO shows an impressive compression ratio of 12.37$\times$ and require storage size of 0.33 MB after compressing eight heterogeneous networks. It is worth noting that we included a new network architecture that YONO did not learn during its offline codebook learning phase. Yet, YONO successfully compress different architectures with an even higher compression rate (11.57$\times$ in \S\ref{subsec:two modality} and 11.77$\times$ in \S\ref{subsec:four modality}) without loss of accuracy on all the unseen datasets.

\textit{In summary, the results here hint that YONO can effectively compress different heterogeneous models trained on unseen datasets without losing accuracy \yd{and demonstrate the generalizability of YONO's codebooks and the effectiveness of the proposed network optimization and optimization heuristics.}}

\subsection{Evaluation on In-Memory Execution and Model Swapping Framework on MCUs}\label{subsec:online eval}

\begin{figure}[t]
  \centering
  \subfloat[Execution]{
    \includegraphics[width=0.224\textwidth]{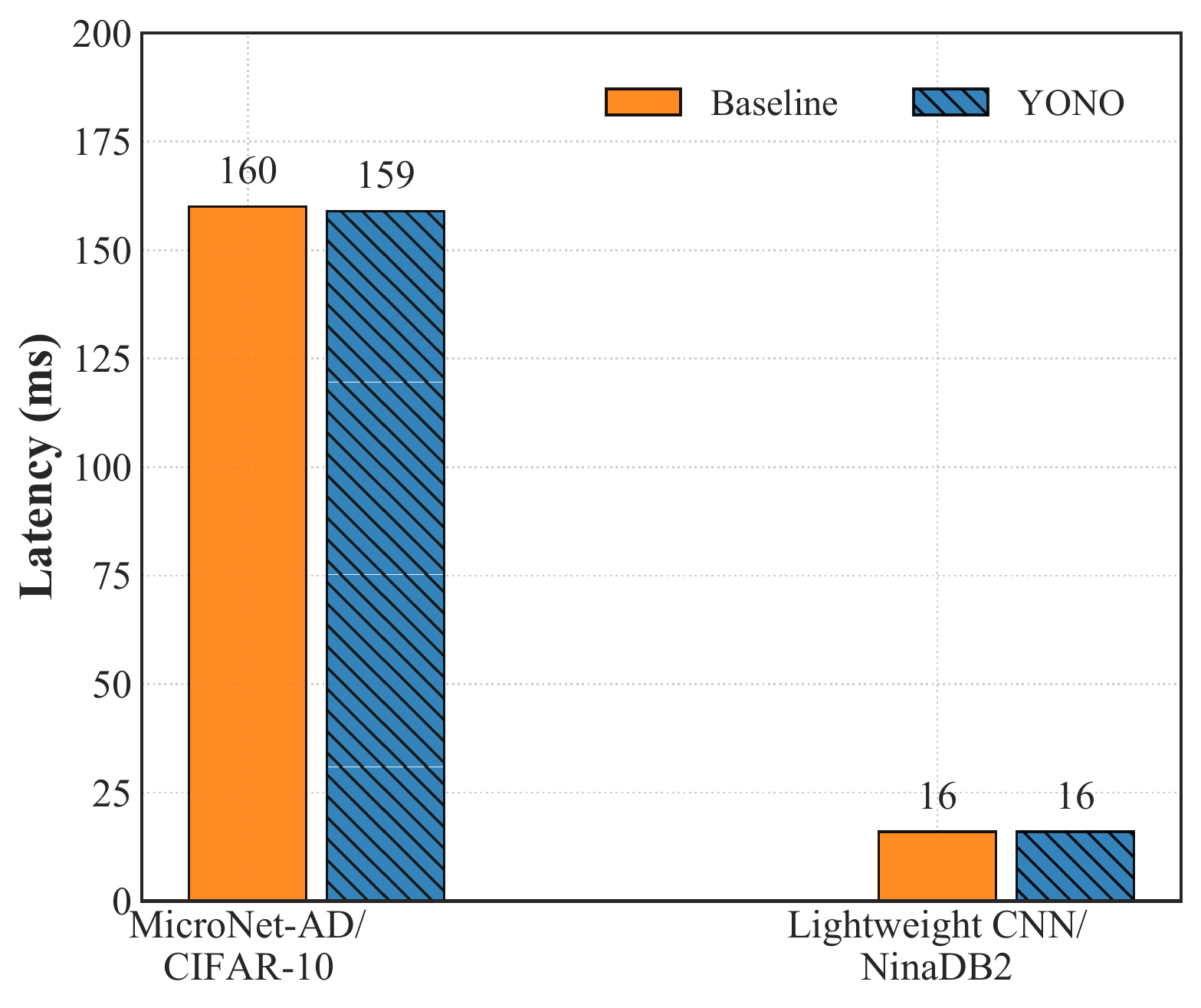}
  \label{fig:execution}
  }
  \subfloat[Loading/Switching]{
    \includegraphics[width=0.224\textwidth]{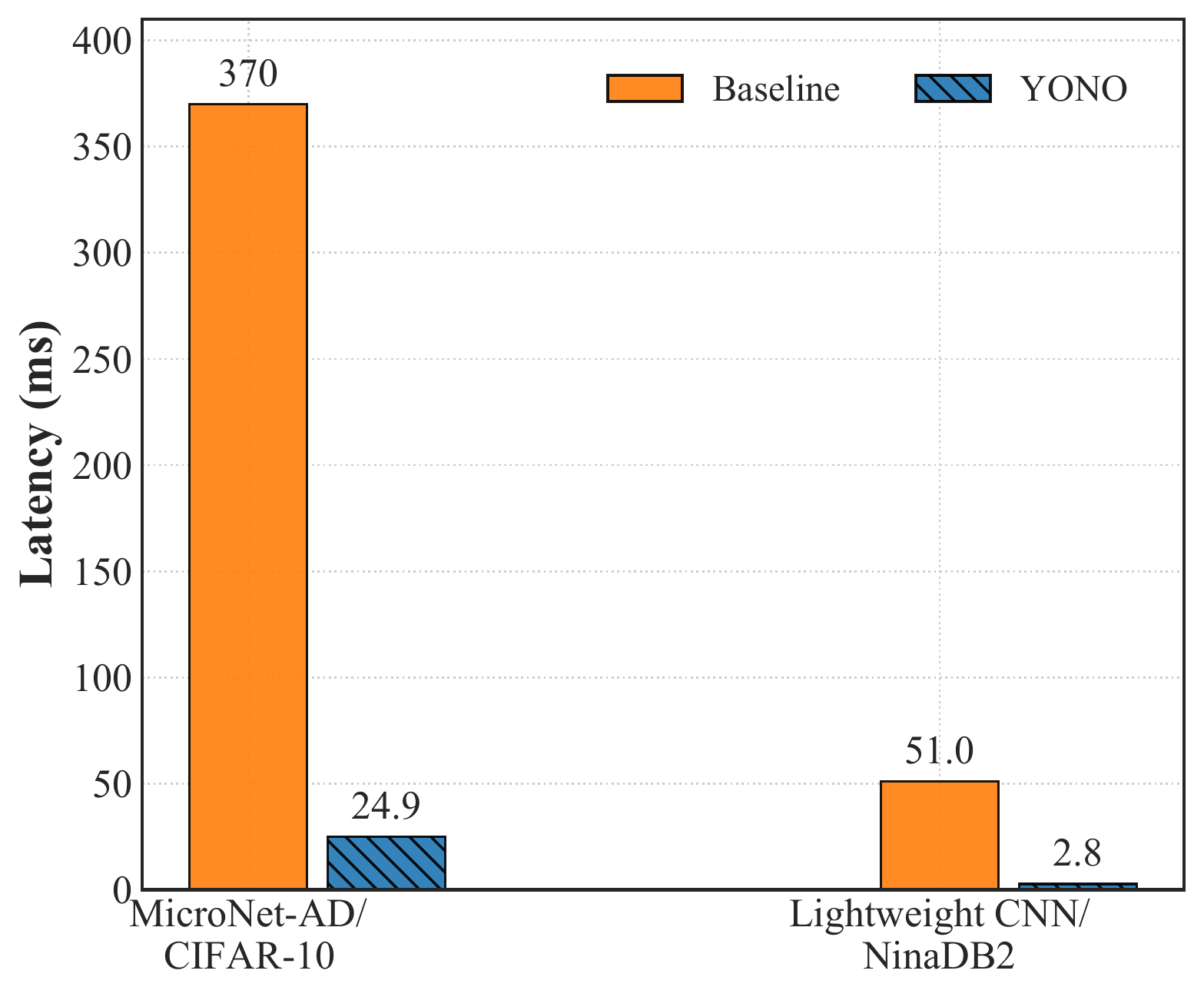}
  \label{fig:switch}
  }
  \caption{
  The model execution and loading/switching time of YONO and the baseline.
  }
  \label{fig:latency}
  \vspace{-0.5cm}
\end{figure}

\begin{figure}[t]
  \centering
  \subfloat[Execution]{
    \includegraphics[width=0.224\textwidth]{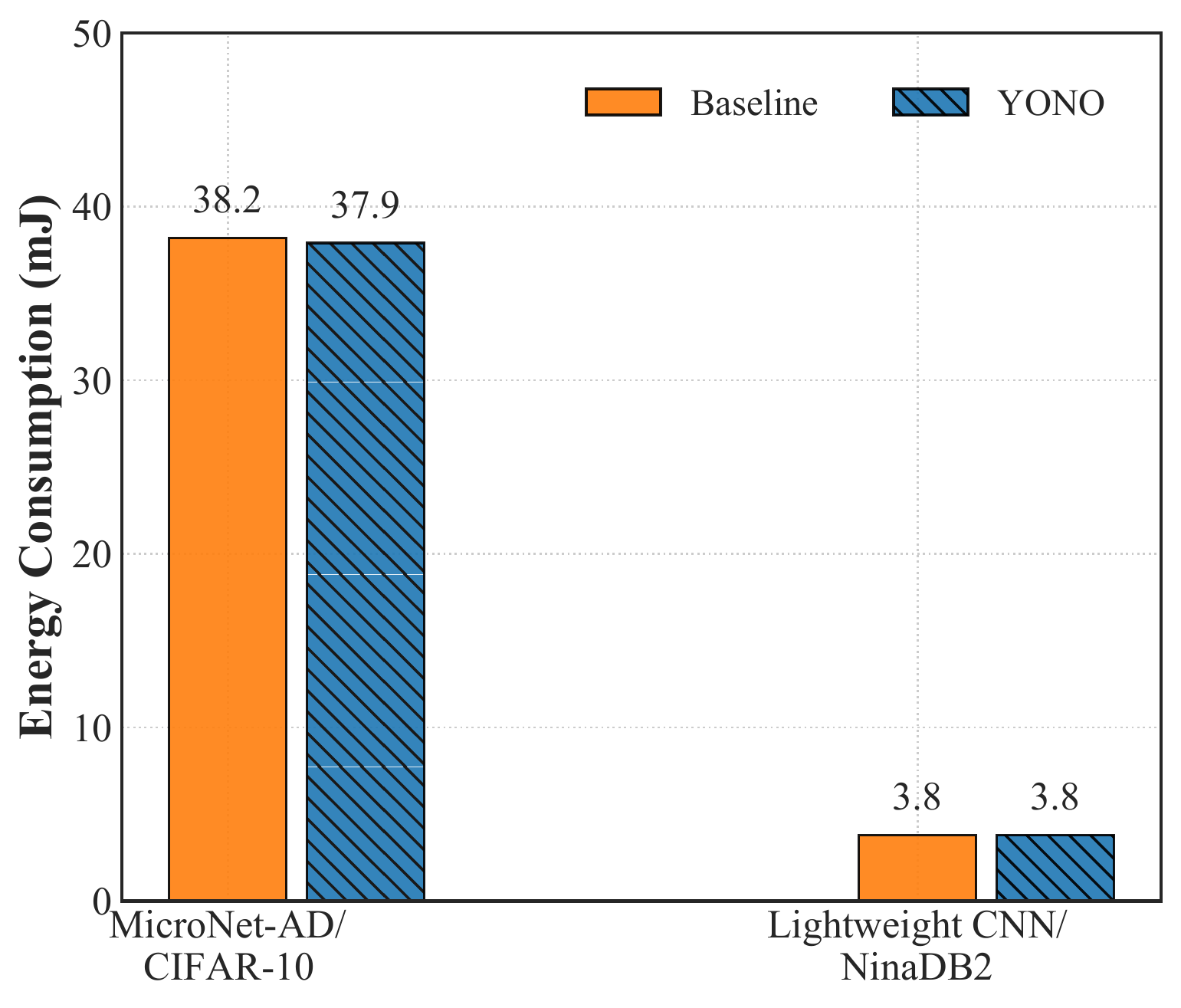}
  \label{fig:energy_execution}
  }
  \subfloat[Loading/Switching]{
    \includegraphics[width=0.224\textwidth]{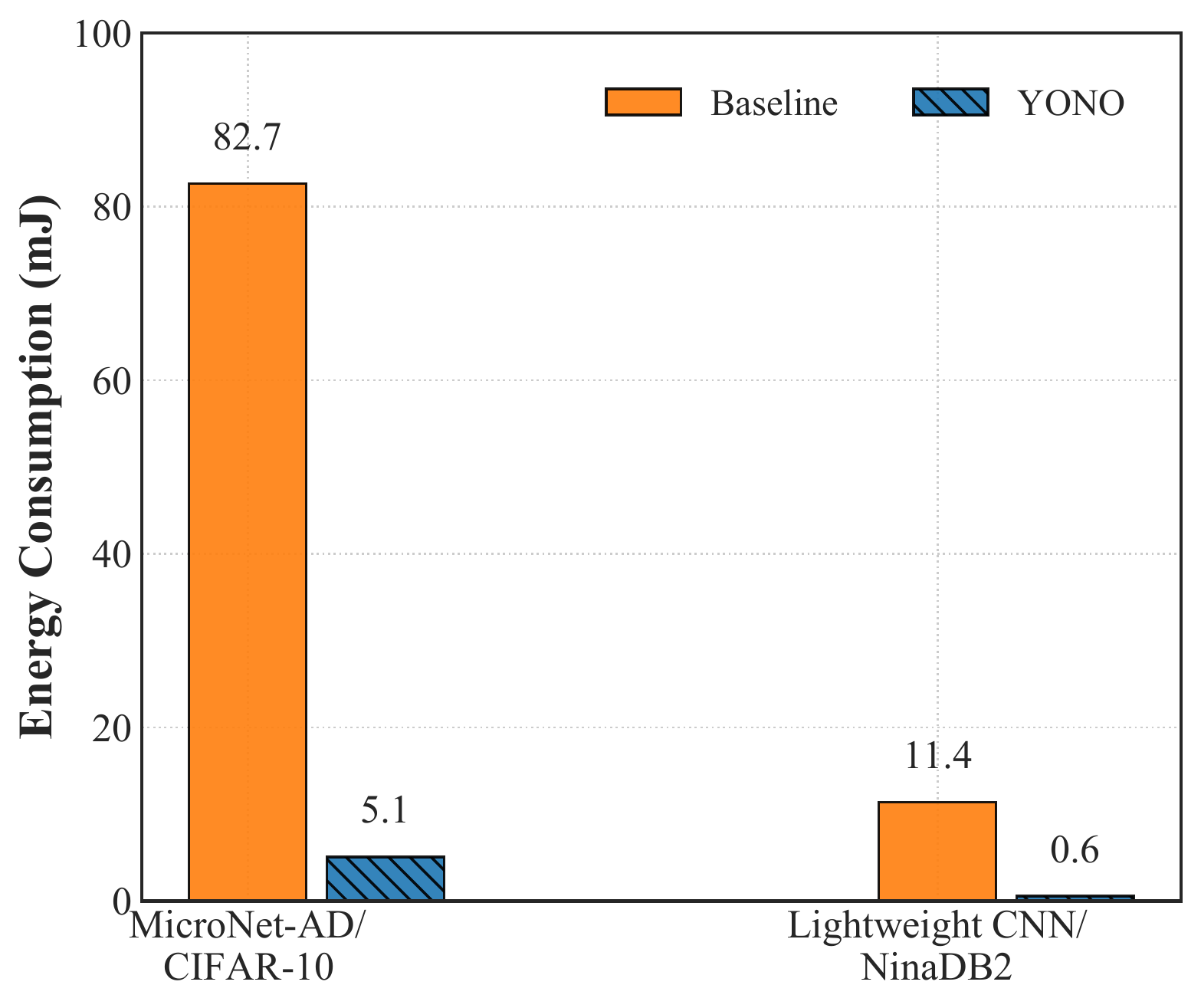}
  \label{fig:energy_switch}
  }
  \caption{
  The energy consumption of model execution and loading/switching of YONO and the baseline.
  }
  \label{fig:energy}
  \vspace{-0.3cm}
\end{figure}

We finally examine the run-time performance of the online component of YONO, the in-memory execution and model swapping framework, introduced in \S\ref{subsec:in_memory}. In specific, we evaluate the latency and energy consumption of model execution and model swapping of YONO on an MCU. \yd{Also, we include an alternative approach to YONO as a baseline that relies on an external SD card as a secondary storage device for storing heterogeneous networks and on in-memory execution similar to YONO.} We employ the same datasets used in the previous subsections. \yd{In Figures~\ref{fig:latency} and \ref{fig:energy}, we report the results of upper bound (i.e., slowest or the most energy-consuming) and lower bound (i.e., fastest or the least energy-consuming) to show the range of latency and energy consumption of YONO and the baseline based on the identified network architectures trained on the datasets in \S\ref{subsec:two modality}-\S\ref{subsec:unseen data} (see Tables~\ref{tab:data_model_arch_1} and \ref{tab:data_model_arch_2}). We use a MicroNet-AD model based on CIFAR-10 as upper bound and a lightweight CNN model based on Ninapro DB2 as lower bound. Although results for other models and datasets are omitted, they reside within the reported latency and energy consumption as in  Figures~\ref{fig:latency} and \ref{fig:energy}.}




\textbf{Latency.} 
We measure the latency of the model execution and model loading/swap by using MBed Timer API, as shown in Figure~\ref{fig:latency}.
In terms of execution time, both YONO and the baseline show a swift execution time (16-160 ms per inference) that can be useful in practice, and there is no meaningful latency difference between them since both rely on in-memory execution. However, for model loading/swap time, YONO accelerates the model switching. YONO reduces model loading/swap time by 93.3\% (370 ms vs. 24.9 ms) in a MicroNet-AD model based on CIFAR-10 and 94.5\% (51.0 ms vs. 2.8 ms) in a lightweight CNN model based on Ninapro DB2 compared to the baseline. Note that we did not conduct a direct comparison on-device with the prior work~\cite{lee_fast_2020} since its source code is not shared and the used MCUs for experiments are not the same. 

\textbf{Energy Consumption.}
We measure the energy consumption of model execution and loading/swap on the MCU using YONO and the baseline, as shown in Figure~\ref{fig:energy}. We use the Tenma 72-7720 digital multimeter to measure the power consumption and then compute the energy consumption over time taken for each operation (i.e., inference and model loading).
Similar to the latency result, the energy consumption for executing models does not show the difference as explained above. However, for the model loading/swap task, YONO decreases energy consumption by at minimum 93.9\% (82.7 mJ vs. 5.1 mJ in a MicroNet-AD model on CIFAR-10) and at maximum 95.0\% (11.4 mJ vs. 0.6 mJ in a lightweight CNN model on Ninapro DB2) compared to the baseline.

\textit{To summarize, the results demonstrate that YONO enables fast (low latency) and efficient (low energy footprints) model execution and loading/swap on an extremely resource-limited IoT device, MCU.}

\section{Discussion}\label{sec:discussion}

\textbf{Impact on Heterogeneous MTL Systems.}
YONO represents the first framework that can compress multiple heterogeneous models and be applicable to unseen datasets. Also, YONO ensures negligible or no loss of accuracy in compressing many different models (architecture) on multiple datasets. This is achieved by only one pair of PQ-based codebooks, our novel optimization procedure, and heuristics. 
\yd{Thus, we envisage that YONO could become a practical system to deploy heterogeneous MTL systems on various embedded devices and platforms in many real-world applications in the future. We leave the wide deployment and performance evaluation of YONO on other embedded platforms under real-world application scenarios as future work.}


\textbf{Application Scenario.}
\yd{Let us consider an example of a real-world application. Given an intelligent authentication system for a smart home, the system would need to detect tenants' identification based on images and voice (image classification and voice recognition). Then, the system could take voice commands as inputs from the identified tenant (e.g., keyword spotting). This simple application scenario already needs three different models, which could satisfy the necessity of a heterogeneous MTL system, YONO.}

\textbf{Generalizability of YONO.}
In Section~\ref{sec:evaluation}, we have demonstrated that YONO can incorporate heterogeneous models and datasets (four different modalities) consisting of 15 datasets (i.e., seven datasets for learning codebooks in \S\ref{subsec:four modality} and the other eight unseen datasets in \S\ref{subsec:unseen data}), which shows that YONO is a generalizable framework. 
\yd{Other datasets and network architectures (e.g., LSTMs~\cite{guan_ensembles_2017} and CNNs with large-sized kernels like 5x5 or 7x7) that can be employed and tested on YONO are left as future work.}


\textbf{Limitation.} 
To enable model switching during the runtime, we design YONO to load the model in the main memory instead of the storage of an MCU. However, since SRAM is a limited on-chip resource and typically smaller than eFlash, our design choice may limit the applicability of YONO, especially for low-end MCUs with smaller SRAM sizes such as 128 KB. Therefore, it would be worthwhile to further investigate memory-efficient ways to reduce the required main memory space for model execution while enabling the model switching at run time. Better usage of FlatBuffer serialization format to hold model weights can be interesting future work since the weights of a model takes the majority of the space.


\section{Related Work}\label{sec:relatedwork}

\textbf{Multitask Learning.}
Multi-task learning allows learning correlated
tasks such that accuracy of both or one of the tasks is improved by exploiting the similarities and differences across tasks~\cite{caruana_multitask_1997}. Common approaches include common feature learning~\cite{liu2015multi,misra2016cross}, low-rank parameter search~\cite{han2016multi,mcdonald2014spectral}, task clustering~\cite{han2015learning,kang2011learning}, and task relation learning~\cite{lee2016asymmetric,long2015learning}. These works achieve limited compression by sharing the first few network layers. However, their main goal is to increase the robustness and generalization of multiple task learners. Thus, keeping multiple heterogeneous DNN models into the extremely limited memory of embedded devices, along with managing and executing
these models (achieving different tasks) efficiently at run-time, are challenging to the aforementioned works. Comparing this, YONO allows to run multiple DNN models efficiently while remaining within the limited resource constraints on embedded devices.

Besides, NWV~\cite{lee_fast_2020} was introduced to compress multiple heterogeneous models of different network architectures and tasks. NWV also minimizes the context switching overhead by retaining all shared weights on the memory. However, NWV's compression ratio is constrained to 8.08$\times$, limiting the multi-tasking IoT system with a small memory footprint to operate many tasks in real-time. Also, the work only employs a simplified LeNet architecture in the experiments of IoT use cases, and thus the accuracy of the system is limited. Conversely, YONO not only increases compression rates but compresses even the highly optimized models (e.g., MicroNet, DS-CNN), while achieving high accuracy that is useful in practice.


\textbf{Mobile and Embedded Sensing Applications.}
Deep learning is increasingly being applied in mobile and embedded systems as it achieves state-of-the-art performances on many sensing applications such as computer vision applications~\cite{fang_nestdnn:_2018}, audio sensing~\cite{lane_deepear:_2015}, activity recognition~\cite{hhar}, gesture recognition~\cite{ninadb2_ninadb3}.
First of all, there exist many vision applications, to name a few, tiny image classification~\cite{cifar10,svhn}, traffic sign recognition~\cite{gtsrb}. 
Besides, audio sensing application is also one of the foundational mobile sensing applications~\cite{kwon_exploring_sec21,servia-rodriguez_knowing_arxiv21} that much research has focused on to deliver behavioral insights to users. The audio sensing tasks include Emotion Recognition (ER)~\cite{emotionsense}, Speaker Identification~\cite{lu_speakersense_2011}, Environmental Sound Classification (ESC)~\cite{su_environment_2019}, and Conversation Analysis~\cite{lee_sociophone_2013}, and Keyword Spotting (KWS)~\cite{zhang_hello_2017}. Next, one of the most widely studied mobile sensing application is HAR~\cite{guan_ensembles_2017,wang_deep_2019}, where the aim is to determine various human activities automatically using body-worn IMU (Inertial Movement Units) sensors.
In application frequently used in mobile sensing is to recognize hand gestures (e.g., fist and open palm) using sEMG (surface Electromyography) signals generated during muscle contractions~\cite{fan_what_2018,becker_touchsense:_2018}. sEMG signal is used for medical~\cite{yousefi_characterizing_2014}, rehabilitation~\cite{winslow_mobile_2018}, human-computer interactions~\cite{kwon_myokey_2020,shatilov_using_2019}, upper-limb prostheses control~\cite{scheme_electromyogram_2011}, and authentication~\cite{chauhan_contauth_imwut20}.

\textbf{Model Compression.}
Many researchers focus on developing a method to improve efficiency without sacrificing the model's accuracy due to a large burden of training deep network architecture and its data~\cite{wu_quantized_2016}. First of all, many researchers have focused on designing and hand-drafting more efficient network architectures, namely, SqueezeNets~\cite{gholami_squeezenext_2018}, ShuffleNets~\cite{ma_shufflenet_2018}, and MobileNets~\cite{howard_mobilenets_2017,sandler_mobilenetv2_2018}, and MicroNet~\cite{banbury_micronets_2021}. In particular, we employ MicroNet as one of our backbone network architectures since it shows impressive performance and efficiency on tiny IoT systems such as MCUs.


In addition, another thread of research is weight pruning methods that leverage the inherent redundancy in the weights of neural networks~\cite{han_deep_2016,yang_designing_2017,mallya2018packnet,liu_autocompress_2020,li_pruning_2016,zhang_systematic_2018}. 
Furthermore, quantization of model weights and activiations has been an active area of research. Many prior works quantize the weights and activations from 32-bit float to 8-bit integer~\cite{jacob_quantization_2018}, ternary values (2-bit)~\cite{zhu_trained_2016,li_ternary_2016}, binary values (1-bit)~\cite{courbariaux_binaryconnect_2015,courbariaux_binarized_2016,rastegari_xnor-net_2016,alizadeh_empirical_2018}, and mixed precision~\cite{wan_tbn_2018,yin_xnor-sram_2020}. Also, weight clustering methods are proposed to group weights into several clusters to compress a model. 

Moreover, researchers studied techniques that quantize an array of scalars of the weights to compress a model or a particular layer. Some works extended a sparse coding~\cite{zhao_review_2016} to learn a compact representation that covers the feature space of weights of a model~\cite{ge_product_2014,bagherinezhad_lcnn_2017}. 
Also, many researchers examined vector quantization-based methods~\cite{wu_quantized_2016}. For example, Gong et al.~\cite{gong_compressing_2014} conducted an empirical study to compare binarized networks, scalar quantization using k-means (i.e., weight clustering), Product Quantization (PQ)~\cite{jegou_product_2011}. Several recent works apply PQ to compress a deep neural network with more than 11 million parameters~\cite{stock_and_2019,stock_training_2020,martinez_permute_2021}. 
Albeit its impressive results, all the prior works only focused on utilizing PQ to a single and bulky model at a scale of millions of parameters, lacking the understanding of how the method can be used to deal with heterogeneous MTL applications and compress tiny models that should fit into the extremely limited memory budget of MCUs (less than 512 KB).  
Thus, for the first time in this work, we develop YONO, a PQ-based model compression framework that operates heterogeneous models on tiny IoT devices. We propose a novel network optimization procedure and heuristics to achieve high accuracy close to the uncompressed models. Also, YONO enables fast and efficient model execution and swapping on an MCU.

\section{Conclusions}\label{sec:conclusions}

We have presented an efficient MTL system, YONO, that compresses multiple heterogeneous models through PQ codebooks, our novel network optimization and heuristics. First, we implemented YONO's offline component on a server and its online component on a critically resource-constrained MCU. Then, we demonstrated its effectiveness and efficiency. YONO compresses multiple heterogeneous models up to 12.37$\times$ with minimal or near to no accuracy loss. Interestingly, YONO can successfully compress models trained with datasets unseen during its offline codebook learning phase. Finally, YONO's online component enables an efficient in-memory model execution and loading/swap with low latency and energy footprints on an MCU. We envision that methods developed for YONO and our research findings could pave the way to deploy practical heterogeneous multi-task deep learning systems on various embedded devices in the near future.


\begin{acks}
This work is supported by a Google Faculty Award, ERC through Project 833296 (EAR), and Nokia Bell Labs through their donation for the Centre of Mobile, Wearable Systems and Augmented Intelligence to the University of Cambridge. 
\end{acks}

\bibliographystyle{ACM-Reference-Format}
\bibliography{ref}

\end{document}